\documentclass[conference]{IEEEtran}
\usepackage{times}
\usepackage{graphicx}

\usepackage[numbers]{natbib}
\usepackage{multicol}
\usepackage[bookmarks=true]{hyperref}

\usepackage{amssymb}
\usepackage{textcomp}
\usepackage{booktabs}
\usepackage{amsmath}
\usepackage{algorithm}
\usepackage{algpseudocode}
\usepackage{multirow}
\usepackage{eso-pic}

\usepackage{todonotes}

\newcommand{\firstpageworkshopnote}{%
  \AddToShipoutPictureFG*{%
    \AtPageLowerLeft{%
      \raisebox{0.35in}{%
        \makebox[\paperwidth]{\footnotesize RSS 2026 Workshop on Diffusion for Robot Learning}%
      }%
    }%
  }%
}

\title{ConFlow: Constraints-Guided Learning with Flow Matching for Motion Generation}

\begin{document}

\author{
Nutan Chen$^{1,\ast}$,
Jianxiang Feng$^{2,\ast}$,
Marvin Alles$^{3}$,
Botond Cseke$^{4}$\\[0.5em]
$^{1}$LS Wiiri Robot Innovation Center, \quad
$^{2}$Independent Researcher,\\
$^{3}$Technical University of Munich,\quad
$^{4}$Volkswagen Group
}

\firstpageworkshopnote
\maketitle

\begingroup
\renewcommand\thefootnote{\textasteriskcentered}
\footnotetext{Equal contribution.}
\endgroup

\begin{abstract}

In recent years Flow Matching has become a prominent method for generative modeling robot motion generation.
In its generic form Flow Matching is an ODE-based neural sampler that is trained by regressing empirical flow fields associated with motion samples as data. However, in robot motion generation we often have additional constraints that might not be present in the collected data.
The majority of current approaches train the flow on the available data and use inference-time guidance to enforce task-specific constraints. To address this mismatch, we propose \textbf{ConFlow}, a constraint-guided flow matching framework that incorporates constraint information directly into the training objective via differentiable barrier or cost functions. To address design specifications such as smoothness and boundary conditions, we propose replacing the standard Gaussian source distribution used in flow matching training with a conditional Gaussian Process. 
Our approach also uses infeasible demonstrations as negative supervision, improving constraint satisfaction without requiring additional expert data. 
Experiments on a two-robot navigation task demonstrate that ConFlow achieves lower collision rates and higher trajectory quality than standard flow matching baselines, with or without inference-time guidance. These results validate training-time constraint integration as an effective approach to closing the training--inference gap in generative motion models.

\end{abstract}

\IEEEpeerreviewmaketitle

\section{Introduction}
Recent advances in generative modeling, particularly diffusion models and flow matching, have driven interest in learning-based robot motion generation. Beyond their ability to model complex motion distributions, these methods also support inference-time guidance, which allows the policy to adapt to new tasks without retraining. This property is especially attractive for robotic deployment in complex and unstructured environments, where guidance can dynamically enforce constraints such as obstacle avoidance~\cite{li2025language} while preserving the expressiveness of the learned motion generator.

A key limitation of current guidance-based motion generation frameworks is the misalignment between the objective optimized during training and the constraints enforced at deployment. While guidance can steer generated motions toward desired behaviors, it effectively modifies the sampling process without altering the underlying learned distribution, resulting in infeasible or physically implausible out-of-distribution behaviors \cite{christopher2026constraint,feng2023topology, guo2024gradient}. %

Furthermore, modern generative models typically require large amounts of training data. In robotics, such models are commonly trained solely on expert demonstrations, whose collection is costly and labor-intensive. This dependence on expert data not only limits scalability but also overlooks abundant non-expert interactions that could provide valuable supervision.
In particular, negative demonstrations provide contrastive information about constraint violations, helping the model distinguish feasible motions from infeasible ones.
Current approaches, including improved inference procedures \cite{wang2025inference} and post-hoc fine-tuning, only partially address these shortcomings. Inference-time guidance cannot fundamentally eliminate the aforementioned discrepancy, whereas fine-tuning often converges to suboptimal solutions and incurs additional training overhead \cite{guo2024gradient}.

To address these limitations, we propose \emph{ConFlow}, a constraint-guided flow matching framework for robot motion generation. Rather than enforcing constraints through inference-time guidance, ConFlow combines (i) a constraint-guided training objective that encourages the learned flow to satisfy task and environmental requirements such as collision avoidance and (ii) a conditional Gaussian Process (GP)-based source distribution that can enforce smoothness and  various boundary or via-point constraints. Together, these components enable the model to learn motion distributions that better align with deployment-time requirements, thereby mitigating the training--inference mismatch. The framework can additionally incorporate infeasible demonstrations as negative supervision, improving constraint satisfaction without requiring additional expert data.
Experiments show that ConFlow consistently improves motion generation quality and robustness to distribution shift.

Our main contributions are as follows: (1) We introduce a constraint-guided learning paradigm that transfers inference-time guidance into the training objective, mitigating the training--inference mismatch. (2) We instantiate this paradigm in a flow matching framework, combining a constraint-aware training objective with a GP-based source distribution. (3) We extend the framework to use non-expert demonstrations, improving data efficiency and generalization without additional expert collection. (4) We validate the approach on a multi-robot obstacle avoidance task, showing improved constraint satisfaction and motion quality.

\section{Related work}
\emph{Generative Motion Generation.}
Diffusion-based methods have shown strong performance in generating robot motions and planning trajectories in high-dimensional spaces. Examples include learning smooth cost functions for joint grasp and motion optimization through diffusion models \cite{urain2023se}, motion planning with diffusion-based trajectory generators \cite{carvalho2023motion}, and physics-guided diffusion models for human motion synthesis \cite{yuan2023physdiff}. These methods demonstrate the ability of generative models to represent complex motion distributions, but they typically rely on either expert demonstrations or inference-time correction mechanisms to enforce constraints.

\emph{Non-expert Demonstrations.}
Collecting large-scale expert robot demonstrations is often expensive and difficult to scale. Recent discussions in robotics have emphasized the importance of leveraging broader sources of data, including non-expert demonstrations and human-generated behaviors, to improve scalability and accessibility of robot learning \cite{billard2025roadmap}.
\citet{knaust2021guided} address this challenge by using Probabilistic Movement Primitives (ProMPs) to capture natural variability in non-expert demonstrations, avoiding the need for carefully crafted expert data.

\emph{Inference-time Guidance and Constraint Satisfaction.}
Inference-time guidance has emerged as an effective way to incorporate constraints and preferences into generative models without retraining. In robot motion generation, guided decoding has been used to adapt robot trajectories online to obstacles and user-specified waypoints \cite{chen2024guided}. Similar ideas have been applied to diffusion policies for collision-aware manipulation \cite{li2025language}, human-intent alignment \cite{wang2025inference}. While effective, these approaches introduce a mismatch between the training objective and the constraint satisfaction during sampling, and often require repeated constraint evaluations and gradient computations during inference~\cite{ christopher2026constraint, guo2024gradient}.

\emph{Training-time Guidance.}
Close to our work, a constraint-aware training framework \cite{christopher2026constraint} employs sequential quadratic programming to project samples into a feasible set, which imposes additional compute consumption during training. Similarly in reinforcement learning, FlowQ \cite{alles2025flowq} introduces energy guidance into flow-matching training by learning a flow that matches an energy-reweighted policy distribution, guiding actions with higher expected return. Additionally, we also share a similar idea of enhancing the expressivity and structural properties of the source distribution with another work in training normalizing flows (NFs)~\cite{feng2023topology}. Our method focuses on robot motion generation and is able to incorporate more constraints.

\section{Preliminaries}
\noindent
\emph{Flow Matching} (FM) \cite{lipman2024flow} is a recently proposed generative modeling framework that learns a continuous-time vector field (flow) to transport samples from a simple source distribution to a target data distribution. Compared with diffusion models, FM directly regresses the target velocity field and therefore avoids the need to simulate and invert a stochastic diffusion process. This formulation often leads to simpler training objectives and faster inference.

Let $x_0 \sim p_0$ denote a sample from a source distribution, typically a standard Gaussian, and $x_1 \sim p_{\mathrm{data}}$ denotes a motion trajectory from the dataset. FM defines a probability path connecting $p_0$ and $p_1$. An intermediate state on this path is given by $x_t = \alpha_t x_1 + \sigma_t x_0$,
where $t\in[0,1]$, $\alpha_t=t$, and $\sigma_t=1-t$. The corresponding ground-truth velocity along this probability path $\psi_t(x_t)$ is $u_t(x_t|x_0, x_1) = \frac{d x_t}{dt}
= \dot{\alpha}_t x_1 + \dot{\sigma}_t x_0.$
A neural network $u_\theta(x,t)$ is trained to approximate this velocity field by minimizing
\begin{equation}
\mathcal{L}_{\mathrm{FM}}(\theta)
=
\mathbb{E}_{t \sim \mathcal{U}_{[0,1]},x_0, x_1}
\left[
\left|
u_\theta(x_t,t)-u_t(x_t|x_0, x_1)
\right|^2
\right].    
\label{eq:fm}
\end{equation}

\noindent
After training, new trajectories can be generated by integrating the learned ordinary differential equation (ODE)
$
\frac{d}{dt}\psi(x_t) = 
u_\theta(x_t,t)
$
starting from a random sample $x_0 \sim p_0$. Numerical solvers such as Euler integration can then transport the sample toward the target motion manifold. %
FM is particularly attractive for robot motion generation because trajectories are naturally continuous and can be represented as points in a high-dimensional trajectory space.

\emph{Gaussian Processes} (GPs)~\cite{williams1995gaussian}, as one of the
popular methods in Bayesian non-parametric modeling, provide a principled
probabilistic framework for performing inference in function space and are thus
widely used in motion / trajectory modelling.
Formally, a GP is
defined as a collection of random variables $\{f_s\}_{s\in  \mathcal{S}}, f_s \in \mathbb{R}^{d}$ indexed by an index set $\mathcal{S}$.
Any finite combination of these random variables $\{f_{s_i}\}_{i=1, \ldots, n}$ indexed by a finite set $\{s_1, \ldots, s_n\}$ follows a joint Gaussian
distribution, specified by a mean function $m_{\phi}(\cdot)$ and covariance function
$k_{\phi}(\cdot,\cdot)$, that is, $\{x_{t_i}\}_{t=1, \ldots, n} \sim \mathcal{N}(\{m_{\phi}(t_i)\}_{i=1, \ldots, n},  \{k_{\phi}(t_i, t_{i'}) \}_{i,i' =1, \ldots, n})$.

Usually, time $\mathcal{S}=[0,T]$ is used as index set. The trajectory values $f_s$ can be Cartesian coordinates or joint angles.
Note that we use $s$ as the time coordinate of the trajectory and $t$ as the time parameter of the flow. For example, $x_t^{s}$ is the $t$ stage intermediate value of the trajectory at time $s$.

Due to the fact that the Gaussian family of distributions is closed under marginalisation and conditioning Bayesian inference and conditioning in these models is analytically tractable.
Given an observation process $x_s \sim  \mathcal{N}(f_s, \sigma^2)$ and observations $\{x_{s_i}\}_{i=1, \ldots, M}$ at $S= \{s_i\}_{i=1, \ldots, M}$, the posterior Gaussian process is defined by the mean and covariance functions
\begin{align}
m_{\phi}(s;  \{x_{s_i}\}) &= m_{\phi}(s)+ \sum_i k_{\phi}(s, s_i)\alpha_i,
\label{eq:gp_posterior}
\\
{k}_{\phi}(s, s'; \{x_{s_i}\}) &= k_{\phi}(s, s') - \sum_{i, i'} k_{\phi}(s, s_i) \Lambda_{i,i'} k_{\phi}(s_{i'}, s'),
\nonumber
\end{align}
with $K = \{k_{\phi}(t_i, t_{i'})\}_{i,i'=1, \ldots, n},  \Lambda= (K+\sigma^2 I)^{-1}, \alpha = \Lambda \{x_{s_i}\}_{i=1,\ldots, M}$.
For conditioning we can set $\sigma^2=0$.
This allows is to generate posterior or conditioned trajectories given some data such as boundary values $x_{s=0}, x_{s=T}$ or via-points.
The parameter $\phi$ can be fitted via max-likelihood to pre-tune the smoothness.
In the following we use posterior or conditioned Gaussian process samples as the source distribution $p_0$.

\section{Method}

\subsection{Problem Statement}
\label{sec:problem}
\noindent
We consider constraint-aware motion generation between prescribed start and goal states. A trajectory is represented as
$[q(s_0), q(s_1), \ldots, q(s_{T-1})]
\in \mathbb{R}^{T \times D}$,
where $q(s_i)\in\mathbb{R}^{D}$ denotes the robot state at normalized trajectory parameter $s_i$. The temporal parameterization is defined as
$s_i = \frac{i}{T-1}, \quad i \in \{0,\ldots,T-1\}$,
such that $s_i\in[0,1]$. For notational convenience, we write $q_i := q(s_i)$, yielding the equivalent representation
${\xi} = [q_0, q_1, \ldots, q_{T-1}]$.
Given a start state $q_{\mathrm{start}}$ and a goal state $q_{\mathrm{goal}}$, our objective is to learn a conditional trajectory distribution
$p_\theta\left(
{\xi} | q_{\mathrm{start}}, q_{\mathrm{goal}}
\right)$,
that generates smooth, diverse, and feasible trajectories satisfying the prescribed task and motion constraints. 

Additionally, we assume access to a dataset consisting of positive demonstrations $\mathcal{D}^{+}$ and negative demonstrations $\mathcal{D}^{-}$. Positive trajectories satisfy all constraints, including collision avoidance, smoothness, and successful task completion, whereas negative trajectories violate one or more of these requirements. Following the flow-matching formulation, we denote a demonstration trajectory by
$x_1 \equiv {\xi}^{\mathrm{data}},
\quad
x_1 \sim p_{\mathrm{data}}$,
where $p_{\mathrm{data}}$ is the empirical trajectory distribution induced by the training dataset. By using both positive and negative demonstrations, we seek to learn a flow-matching model that assigns high probability to feasible motions while suppressing constraint-violating behaviors, thereby improving constraint satisfaction without sacrificing trajectory diversity.

\subsection{Constraint-Guided Flow Matching}
\noindent
To align training with deployment-time requirements, we introduce two complementary modifications to standard conditional flow matching: (i) a \emph{constraint-guided training objective} that incorporates task constraints directly into the learned velocity field, and (ii) a \emph{constraint-injected source distribution} that injects structural motion priors into the generative process. Together, these components encourage the learned flow to generate trajectories that satisfy motion constraints while preserving trajectory diversity.

\subsubsection{Constraint-Guided Objective}

A key limitation of existing guidance-based motion generation methods is that constraints are enforced only during inference, creating a \textit{mismatch} between the training objective and deployment-time behavior. To mitigate this issue, we incorporate constraint guidance directly into the flow-matching training objective.

Given the standard conditional flow-matching loss in Eq.~\eqref{eq:fm}, we define the constraint-guided objective as

\begin{align}
\mathcal{L}_{\mathrm{CG}} = 
\mathrm{MSE}
\left(
u_t^\theta -
u_t -
\lambda_c M(x_1) u_t^c
\right)
,
\label{eq:cg_loss}
\end{align}
where $u_t^\theta$ denotes the predicted velocity field, $u_t$ is the standard flow-matching target velocity, $MSE$ denotes the mean squared error function and
\begin{align}
u_t^c = \nabla_{x_t} C(x_t)
\end{align}
is a constraint-induced correction derived from a differentiable constraint function $C(\cdot)$. The coefficient $\lambda_c$ controls the strength of the constraint guidance.

To use both expert and non-expert demonstrations, we train on a dataset $\mathcal{D}=$
$\mathcal{D}^{+}
\cup
\mathcal{D}^{-}$,
where $\mathcal{D}^{+}$ contains feasible trajectories and $\mathcal{D}^{-}$ contains trajectories that violate one or more constraints. For each target trajectory $x_1$, we define a binary violation indicator
\begin{align}
M(x_1) = 
\begin{cases}
1, & \text{if } x_1 \text{ violates the constraint},\\
0, & \text{otherwise}.
\end{cases}
\end{align}
The correction term is therefore activated only for constraint-violating demonstrations. This design prevents over-regularization of feasible trajectories while enabling negative demonstrations to provide informative supervision about undesirable behaviors. 
Specifically, we instantiate the framework using a collision-avoidance constraint. While prior work commonly employs ReLU-based collision penalties, we instead adopt a SoftPlus formulation to obtain smoother gradients:
\begin{align}
C_\gamma(x_t)
=
-\frac{1}{\gamma}
\log\left(
1+\exp\left(
-\gamma\bigl(d(x_t)-r\bigr)
\right)
\right).
\label{eq:c}
\end{align}
where $d(x_t)$ denotes the minimum distance between two robots (or robot links) and $r$ is the corresponding safety radius. $\gamma>0$ is the softplus sharpness
parameter. A larger $\gamma$ yields a sharper transition around the
safety boundary $d(x_t)=r$, whereas a smaller $\gamma$ provides a
smoother gradient over a wider neighborhood of the boundary. The resulting gradient $u_t^c$ pushes trajectories away from collision configurations and serves as the constraint-guided correction during training.

\subsubsection{Constraint-Injected Source Distribution}

In addition to the proposed training objective, we incorporate motion constraints directly into the source distribution of flow matching. A key observation is that the standard Gaussian source distribution assumes \textit{independent} noise across trajectory waypoints, potentially leading to arbitrarily varying neighboring states.

To address this limitation, we inject trajectory smoothness into the source distribution by replacing the standard Gaussian prior with a GP. This design follows a similar intuition to the multimodal source distributions used to alleviate topology mismatch in generative transport models~\cite{feng2023topology}. Rather than learning smoothness solely through the flow, we encode it directly into the initial trajectory distribution. 

Specifically, instead of sampling $x_0 \sim p_0$, we draw source trajectories from a GP $x_0 \sim p_{\mathrm{GP}}$.
Let ${s_i}$ denote the normalized trajectory parameters introduced in Sec.\ref{sec:problem}. Each trajectory dimension $d$ is modeled independently as
$f_d(s)
\sim
\mathcal{GP}(0,k(s,s'))$,
where $k(\cdot,\cdot)$ is an RBF kernel. Discretizing the process at the trajectory waypoints yields 
${x}_{0,d} = [f_d(s_0),\ldots,f_d(s_{T-1})]^T \sim \mathcal{N}(0,K)$,
with covariance matrix $K_{ij} = \sigma^2
\exp
\left(
-\frac{(s_i-s_j)^2}{2\ell^2}
\right)
+
\epsilon\delta_{ij}$,
where $\ell$ is the kernel length scale and $\epsilon=10^{-6}$ is added for numerical stability. GP samples exhibit strong temporal correlations between neighboring waypoints. 
Consequently, the learned flow is encouraged to focus on modeling meaningful trajectory variations rather than correcting high-frequency artifacts introduced by an \textit{unstructured} Gaussian source distribution.

Furthermore, we optionally condition the GP on the trajectory endpoints. Given a target trajectory $x_1$, we extract its start and goal states,
$y_{\mathrm{start}} = 
x_1(s_0),
\quad
y_{\mathrm{goal}} = 
x_1(s_{T-1})$,
and construct a \textit{conditional GP} source distribution
\begin{align}
x_0
\sim
p_{\mathrm{CGP}}
\Big(
x
\mid
x(s_0)=y_{\mathrm{start}},
x(s_{T-1})=y_{\mathrm{goal}}
\Big).
\end{align}

To enforce boundary conditions, we condition the GP on the start and goal states of the target trajectory. Let
$\mathcal{O}=\{0,T-1\}$
denote the endpoint indices and
$\mathcal{I}=\{1,\ldots,T-2\}$
the interior indices. Partitioning the covariance matrix as
$
K=
\begin{bmatrix}
K_{\mathcal{OO}} & K_{\mathcal{OI}}\\
K_{\mathcal{IO}} & K_{\mathcal{II}}
\end{bmatrix},
$
Given the start endpoint observations of dimension $d$: $y_{\mathcal{O},d}
=
\begin{bmatrix}
y_{\mathrm{start},d} \\
y_{\mathrm{goal},d}
\end{bmatrix}$.
According to Eq.\ref{eq:gp_posterior}, the conditional distribution of the interior trajectory values follows the standard GP posterior

\begin{align}    
\label{eq:GP_source_dist}
x_{0,\mathcal I,d}
\mid
x_{0,\mathcal O,d}=y_{\mathcal O,d}
\sim
\mathcal N
\!\left(
\mu_{\mathcal I|\mathcal O,d},
\Sigma_{\mathcal I|\mathcal O}
\right),
\end{align}
with
$\mu_{\mathcal I|\mathcal O,d}=
K_{\mathcal{IO}}
K_{\mathcal{OO}}^{-1}
y_{\mathcal O,d}$,    
$\Sigma_{\mathcal I|\mathcal O}=
K_{\mathcal{II}}
-
K_{\mathcal{IO}}
K_{\mathcal{OO}}^{-1}
K_{\mathcal{OI}}$.
The full source trajectory is obtained by fixing the endpoints:$x_0(s_0)=y_{\mathrm{start}}$,
$
x_0(s_{T-1})=y_{\mathrm{goal}},
$ and sampling only the interior states from the conditional GP in Eq.\ref{eq:GP_source_dist}.
The resulting flow-matching path remains unchanged: 
$x_t = (1-t)x_0 + t x_1$,
but now $x_0$ and $x_1$ share identical start and goal states. The learned flow therefore focuses on refining the intermediate trajectory while preserving smoothness and endpoint consistency, substantially simplifying the transport problem.

\begin{algorithm}[t]
\caption{Guided Flow-Matching Training}
\label{alg:guided_fm_training}
\begin{algorithmic}[1]

\Require $q(x_1)$, $p_0(x_0)$, $v_\theta$, $\lambda$
\Ensure Trained $v_\theta$

\Repeat
    \State $x_1 \sim q(x_1)$
    \State $x_0 \sim p_0(x_0)$ \hfill $\triangleright$ GP or Gaussian source
    \State $t \sim \mathcal{U}(0,1)$
    \State $x_t = (1-t)x_0 + t x_1$,\quad $\dot{x}_t = x_1 - x_0$ 
    \State $g_t = \nabla_{x_t} C(x_t)$ \hfill $\triangleright$ constraint guidance, Eq.~\eqref{eq:c}
    \State $M(x_1) \in \{0,1\}$ \hfill $\triangleright$ feasibility mask
    \State $u_t = \dot{x}_t + \lambda\, M(x_1)\, g_t$ \hfill $\triangleright$ guided target velocity
    \State Update $\theta$:\;
    $\mathcal{L}(\theta) = \bigl\|v_\theta(x_t,t) - u_t\bigr\|_2^2$
\Until{convergence}
\State \Return $v_\theta$

\end{algorithmic}
\end{algorithm}

\subsection{Inference}
\noindent
Unlike conventional guidance-based approaches that rely heavily on deployment-time corrections, ConFlow learns constraint-aware velocity fields during training. As a result, inference follows the standard flow-matching sampling procedure and requires little or no additional guidance to generate feasible trajectories. Optional guidance can nevertheless be incorporated to further improve safety in challenging scenarios.
\paragraph{Boundary-conditioned initialization}
Consistent with training, sampling is initialized from a conditional Gaussian Process (GP) prior. Given prescribed start and goal states, the source trajectory $x_0$ is sampled from the conditional GP distribution, ensuring that the generated trajectory satisfies the desired boundary conditions from the outset. The learned flow then transports this smooth prior trajectory toward the target motion distribution.

\paragraph{Collision-avoidance guidance}
To further improve safety during deployment, collision avoidance can be incorporated as an inference-time guidance term. Given the learned velocity field $v_\theta(x_t,t)$, we modify the sampling dynamics as
$v_\theta(x_t,t)
+
\lambda \nabla_{x_t} C(x_t)
$,
For robot--robot avoidance, the barrier function $C(\cdot)$ is defined using the minimum distance between the two robots. For obstacle avoidance, the barrier is computed from the minimum distance between each robot and the obstacle. During sampling, the barrier gradient continuously modifies the velocity field, encouraging collision-free trajectories while preserving the motion prior learned during training. In our setup, obstacle avoidance and robot–robot collision avoidance are formulated in essentially the same way. The main difference is the fixed obstacles position with different size to the robots.

\section{Experiments}

\subsection{Task Description}
\begin{figure*}
\centering
\includegraphics[width=0.8\linewidth]{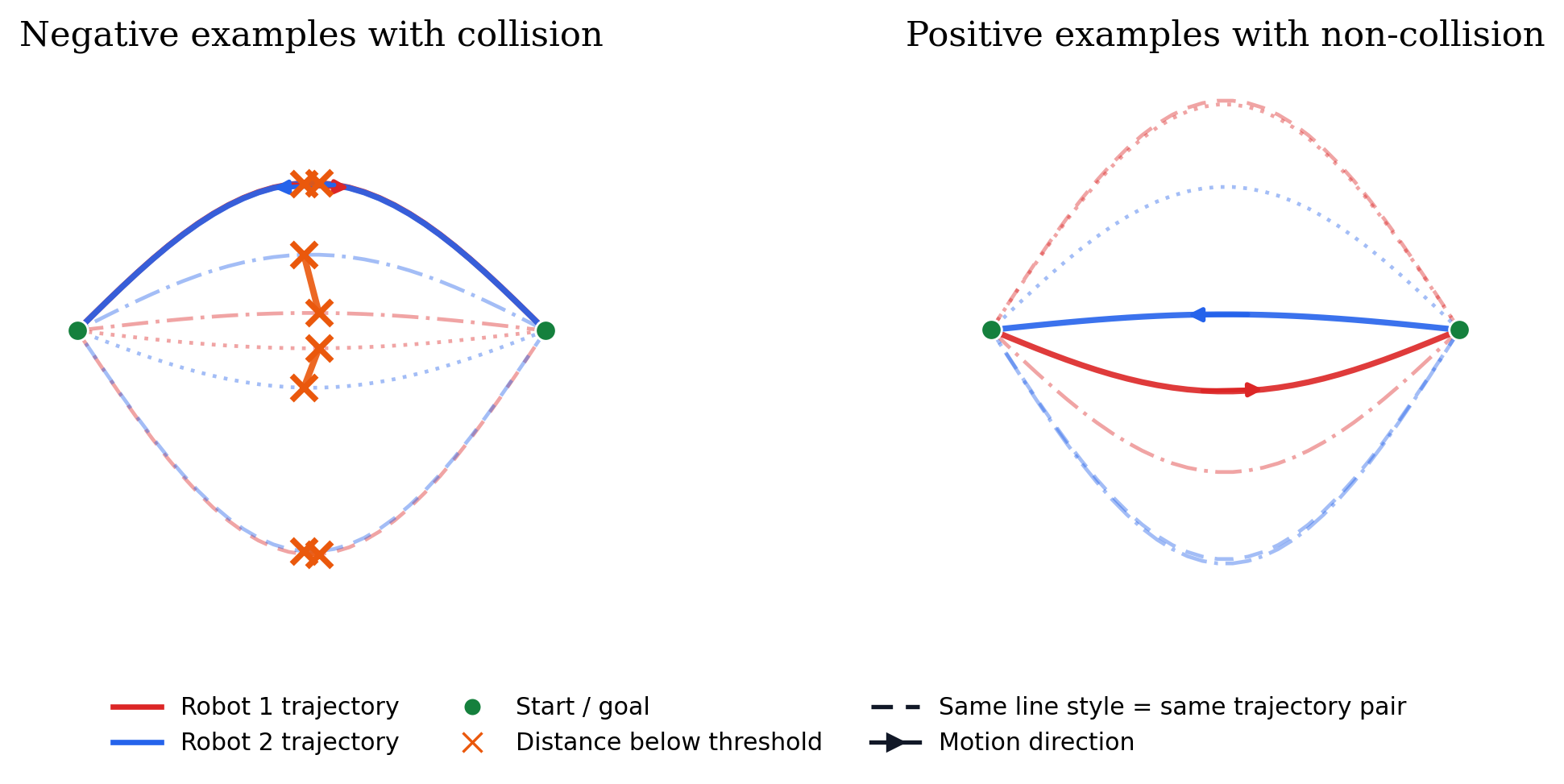}
\caption{Two robots exchanging positions in the synthetic navigation task. The dataset contains both collision-free and collision-inducing trajectories, enabling the evaluation of constraint-guided learning with positive and negative demonstrations.}
\label{fig:placeholder}
\end{figure*}

\noindent
Following the setup of \cite{shaoul2025multirobot}, we consider a two-robot trajectory generation task in which two robots exchange positions in a two-dimensional workspace. Each robot starts from one side of the workspace and moves toward the other's initial position, as illustrated in Fig.~\ref{fig:placeholder}. To generate diverse interaction patterns, we perturb the nominal straight-line motions with randomized sinusoidal deviations orthogonal to the direction of travel, producing smooth trajectories with varying levels of interaction and collision risk.

Trajectories are represented as fixed-length sequences of Cartesian positions and normalized to the range $[-1,1]$ along each dimension. Each trajectory contains $l=32$ waypoints and $n=4$ state dimensions corresponding to the planar coordinates of both robots. Importantly, approximately $30\%$ of the demonstrations contain robot--robot collisions, providing a mixture of feasible and infeasible trajectories for evaluating the proposed constraint-guided learning framework.

\subsection{Evaluation Setup and Metrics}
\paragraph{Evaluation Setup}
For each trained model, we generate 256 trajectories under three inference settings in a single run: (i) robot-avoidance guidance, (ii) obstacle-avoidance guidance, and (iii) no guidance. All ConFlow variants are initialized from the same conditional GP source distribution, conditioned on the start and goal states of evaluation trajectories.

\paragraph{Baselines}
We compare ConFlow against two ablated variants that isolate the contributions of constraint-guided learning and negative demonstrations.

\begin{itemize}
\item \emph{ConFlow w/o Neg. Data}: identical to ConFlow but trained on feasible demonstrations only. This ablation isolates the contribution of infeasible demonstrations as negative supervision.

\item \emph{FM}: standard conditional flow matching trained without constraint-guided learning and with a Gaussian source distribution. This baseline evaluates whether vanilla generative modeling, augmented with inference-time guidance, is sufficient for safe motion generation.

\end{itemize}

Together, these ablations disentangle the effects of (i) the constraint-guided training objective and GP source distribution and (ii) negative demonstrations.

\paragraph{Metrics}
We report two safety-oriented metrics. The \emph{robot-collision rate} measures the fraction of generated trajectories in which the two robots violate a predefined collision threshold. The \emph{object-collision rate} measures the fraction of trajectories in which either robot intersects a circular obstacle introduced during evaluation.
Obstacle avoidance is not enforced during sampling under the Robot-Avoidance and No-Guidance settings; consequently, the corresponding metric is omitted from the evaluation.

\begin{table*}[t]
\centering
\caption{Collision-rate ablation on the two-robot task. Lower is better. Results are reported under three inference \\ settings: no guidance, robot-avoidance guidance, and obstacle-avoidance guidance. Training set configuration is \\ indicated in the bracket: ($\mathcal{D}^{+}$), ($\mathcal{D}^{-}$) denote positive and negative data, respectively.}
\label{tab:two_robot_collision_ablation}
\resizebox{0.8\textwidth}{!}{
\begin{tabular}{l|c|c|cc}
\toprule
&
\multicolumn{1}{c|}{No Guidance}
&
\multicolumn{1}{c|}{Robot Avoidance}
&
\multicolumn{2}{c}{Obstacle Avoidance}
\\
\cmidrule(lr){2-2}
\cmidrule(lr){3-3}
\cmidrule(lr){4-5}
Method
& Robot %
& Robot %
& Robot & Object
\\
\midrule
ConFlow  ($\mathcal{D}^{+}
\cup
\mathcal{D}^{-}$)
& \textbf{0.016} %
& \textbf{0.004} %
& \textbf{0.000} & \textbf{0.035}
\\
ConFlow ($\mathcal{D}^{+}$)
& 0.043 %
& \textbf{0.004} %
& 0.035 & 0.074
\\
FM ($\mathcal{D}^{+}$)
& 0.031 %
& 0.020 %
& 0.039 & 0.039
\\
FM ($\mathcal{D}^{+}
\cup
\mathcal{D}^{-}$)
& 0.055 %
& 0.270 %
& 0.281 & 0.059
\\
\bottomrule
\end{tabular}
} %
\end{table*}

\subsection{Results}
\noindent
Table~\ref{tab:two_robot_collision_ablation} compares ConFlow against its ablations and the Flow Matching (FM) baseline under different inference-guidance settings. Overall, ConFlow consistently achieves the lowest collision rates across all evaluation modes, demonstrating superior generalization and robustness. This validates our central hypothesis that transferring guidance from inference to training can reduce the training--deployment mismatch while improving robustness. 
This also indicates that the benefits of constraint-guided learning are complementary to inference-time guidance. 

The performance gain of ConFlow ($\mathcal{D}^{+} \cup \mathcal{D}^{-}$) over ConFlow ($\mathcal{D}^{+}$) further highlights the importance of utilizing non-expert demonstrations under our proposed framework. In contrast, as expected, naively including negative examples into a standard FM degrades performance due to the ambiguous supervision signal introduced by mixing positive and negative samples.

A key observation is that, even without any inference-time guidance, ConFlow achieves substantially lower robot-collision rates than FM (0.016 vs. 0.031). This result suggests that the collision-avoidance behavior has been internalized by the learned velocity field during training rather than being enforced solely through test-time corrections. Consequently, ConFlow remains robust even when deployment-time guidance is unavailable or imperfect.

\subsection{Ablation on Source Distributions}
\noindent
Table~\ref{tab:source} evaluates the impact of the proposed conditional Gaussian Process (CGP) source distribution. Replacing the standard Gaussian prior with CGP consistently improves all evaluated metrics. In particular, the robot-collision rate is reduced by nearly $46\%$ ($0.0288 \rightarrow 0.0156$), while both median and $95$-th-percentile smoothness metrics improve substantially. Furthermore, endpoint errors decrease by more than an order of magnitude, indicating that conditioning the source distribution on the start and goal states effectively preserves task boundary conditions during generation.

These results highlight two key advantages of the proposed source distribution. First, the GP prior introduces temporal correlations between neighboring waypoints, encouraging smooth trajectory structures before any flow transformation is applied. Second, endpoint conditioning embeds task-relevant boundary information directly into the source distribution, reducing the burden on the learned flow to satisfy start--goal constraints.

\begin{figure}[t]
    \centering
    \includegraphics[width=1.0\linewidth]{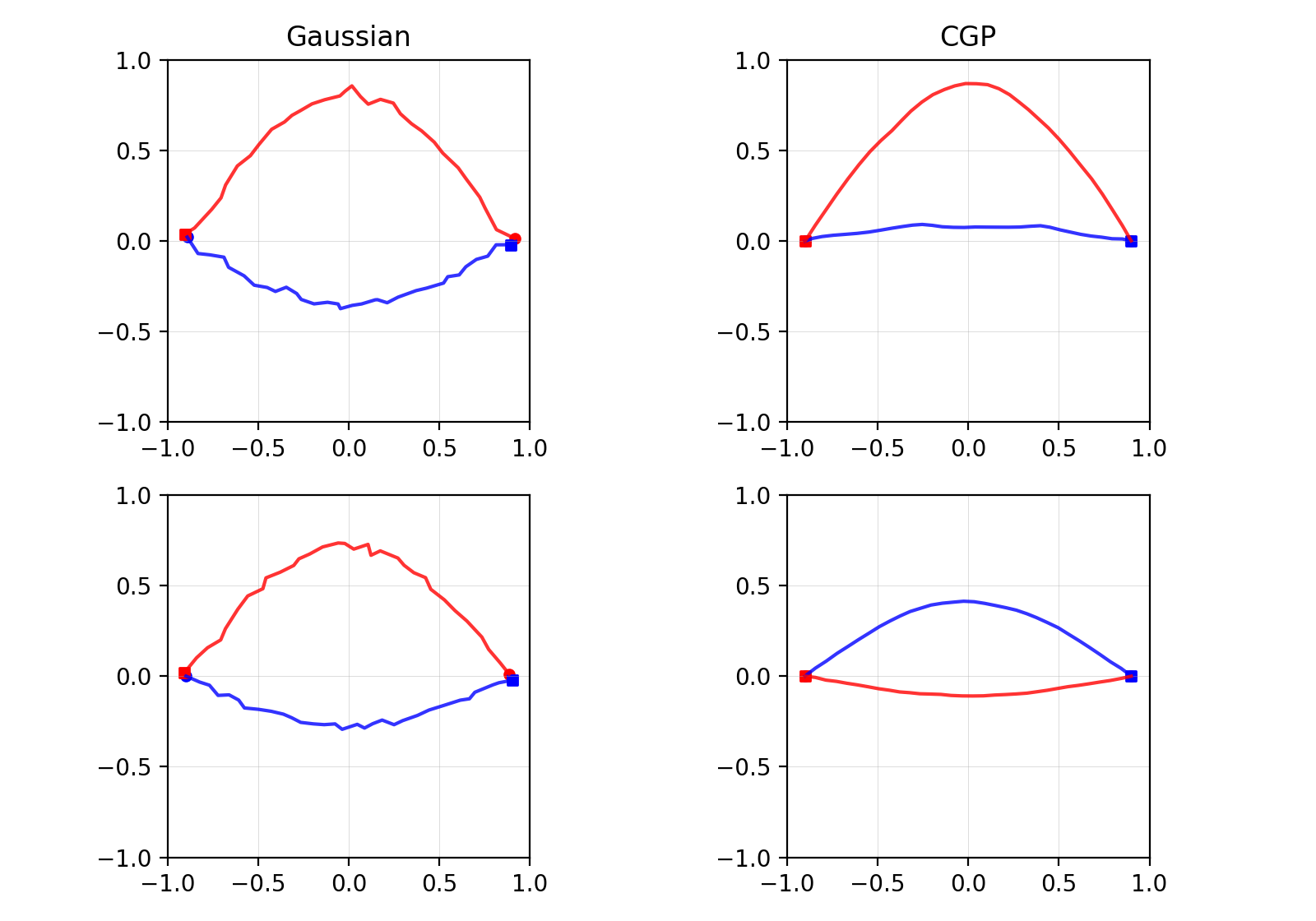}
    \caption{Visualization of generated trajectories using Gaussian and conditional GP (CGP) as source distributions. The Gaussian produces trajectories with more local oscillations, while CGP yields much smoother motions. Moreover, without endpoint conditioning, trajectories from Gaussian drifted far from the data distribution, leading to unstable samples.}
    \label{fig:priors}
\end{figure}

The qualitative results in Fig.~\ref{fig:priors} further support these observations. Compared with a standard Gaussian prior, CGP produces noticeably smoother trajectories with fewer local oscillations. We additionally observed that unconstrained GP priors were sensitive to kernel-scale selection and occasionally generated trajectories that drifted far from the data manifold. Such behavior can lead to unstable samples and degraded performance, particularly in safety-critical motion-generation tasks. Conditioning the GP on the start and goal states effectively mitigates this issue by preserving both temporal structure and task-specific boundary information. Overall, these results demonstrate that incorporating \textit{structured priors} into the source distribution significantly improves the quality and stability of the learned flow.

\begin{table}[t]
\centering
\caption{Ablation Study on Source Distributions}
\label{tab:source}
\resizebox{0.8\columnwidth}{!}{
\begin{tabular}{lcc}
\toprule
Metric & Gaussian & CGP \\
\midrule
Collision rate $\downarrow$ & 0.0288 & \textbf{0.0156} \\
Smoothness median $\downarrow$ & 0.0678 & \textbf{0.0174} \\
Smoothness p95 $\downarrow$ & 0.2218 & \textbf{0.0309} \\
Start error / robot $\downarrow$ & 0.0267 & \textbf{0.0006} \\
End error / robot $\downarrow$ & 0.0276 & \textbf{0.0006} \\
\bottomrule
\end{tabular}
}
\end{table}

\section{Conclusion}
\noindent
We present ConFlow, a constraint-guided flow matching framework for robot motion generation that integrates deployment-time constraints directly into the training objective. By combining a constraint-guided loss with a conditional Gaussian Process source distribution, ConFlow reduces the training--inference mismatch inherent in guidance-based generative models while promoting smooth and safe trajectory generation.
Experiments on a multi-robot navigation benchmark demonstrate improved collision avoidance and motion quality compared with standard flow matching baselines. The framework also benefits from infeasible demonstrations as negative supervision, yielding further reductions in constraint violations without requiring additional expert data.
Future work will extend ConFlow to higher-dimensional robotic systems, such as articulated manipulators, and evaluate performance on real-world platforms. More broadly, these results suggest that jointly designing the training objective and source distribution — rather than treating them as independent components — is an effective approach to embedding task-relevant structure into flow-based generative models, and represents a promising direction for safer and more capable robot motion generation.

\bibliographystyle{plainnat}
\bibliography{example}

\appendix

\paragraph{Architecture and training details.}
We parameterize the velocity field with a residual multilayer perceptron (ResMLP). Each trajectory has length $32$ and state dimension $4$, corresponding to the two planar robot positions $(x_1,y_1,x_2,y_2)$. The trajectory is flattened before being passed to the network, giving an input dimension of $32 \times 4 = 128$. The ResMLP uses a scalar time input, hidden dimension $1024$, and $4$ residual blocks. The model is trained with AdamW using learning rate $3\times 10^{-4}$, batch size $2048$, and $20001$ training iterations. Gradients are clipped with maximum norm $3.0$.

For each training batch, we sample target trajectories $x_1$ from the two-robot training dataset and construct the source trajectory $x_0$ from a conditional Gaussian-process prior. The prior is conditioned on the start and goal states of $x_1$, with length scale $0.4$ and variance parameter $\sigma_f=1.0$. We use an affine probability path with a conditional optimal-transport scheduler. The full training dataset contains $10000$ trajectories and includes both collision-free and robot-collision demonstrations.

The proposed constrained model modifies the standard flow-matching objective with a robot--robot collision-avoidance term. Let $v_\theta(x_t,t)$ be the learned velocity field and $\dot{x}_t$ be the target velocity from the probability path. We compute a softplus collision barrier between the two robots using collision threshold $0.3$ and sharpness parameter $\gamma=10.0$. The gradient of this barrier with respect to $x_t$ is smoothed along the trajectory using a Gaussian filter with $\sigma=5.0$ and kernel size $9$. 
For training,
$\lambda_{\mathrm{c}}=2.5$, $C(x_t)$ is the robot--robot collision barrier, and $M(x_1)$ is a binary mask indicating whether the target trajectory contains a robot--robot collision under threshold $0.3$. Thus, the collision-avoidance correction is applied only to collision-risk training samples.

The ODE solver step size is 0.004.
For inference-time guidance, the robot-collision and obstacle-avoidance guidance weights are set to 0.08 and 1.0, respectively.

\end{document}